\documentclass[letterpaper, 10 pt, journal, twoside]{IEEEtran}
\IEEEoverridecommandlockouts
\usepackage{cite}
\usepackage{amsmath,amssymb,amsfonts}
\usepackage{graphicx}
\usepackage{textcomp}
\usepackage{xcolor}

\usepackage{siunitx}
\usepackage[export]{adjustbox}
\usepackage{setspace} 
\usepackage[font=small]{caption}
\usepackage{subcaption}
 \usepackage{booktabs}
\usepackage{tabularx} 
\usepackage{graphicx} 
\usepackage{listings}
\usepackage{textcomp}
\usepackage{url}
\usepackage{balance}
\usepackage{multirow}
\usepackage{hyperref}
\usepackage{cleveref}
\usepackage{wrapfig}
\usepackage{mathtools}
\crefformat{footnote}{#2\footnotemark[#1]#3}
\usepackage{ragged2e} 
\usepackage[subnum]{cases}
\usepackage{algorithm, algcompatible}
\usepackage{bm}
\usepackage[noend]{algpseudocode}
\usepackage{hyperref}
\usepackage{soul,xcolor}
\usepackage[bottom]{footmisc}
\setstcolor{red}
\usepackage{soul}
\usepackage{amsthm}
\usepackage{tikz}
  
\theoremstyle{definition}

\newcommand{\eg}{{e.g.},}
\newcommand{\ie}{{i.e.},}

\DeclarePairedDelimiterX{\norm}[1]{\lVert}{\rVert}{#1}
\def\secref#1{Sec.~\ref{#1}}
\def\figref#1{Fig.~\ref{#1}}
\def\tabref#1{Table~\ref{#1}}

\def\cblue{\textcolor{black}}

\algnewcommand\algorithmicto{\textbf{to}}
\algnewcommand\RETURN{\State \textbf{return} }

\newcommand{\forestA}{\textit{Forest I}}
\newcommand{\forestB}{\textit{Forest II}}
\newcommand{\forestC}{\textit{Forest III}}
\newcommand{\ourmap}{\text{UA-occupancy}}

\graphicspath{{./figures/}}

\newcommand{\zcal}{\mathcal{Z}}
\newcommand{\fcal}{\mathcal{F}}

\algrenewcommand{\algorithmiccomment}[1]{\hskip2.5em #1}

\definecolor{darkgreen}{rgb}{0,0.5,0}
\definecolor{salmon}{rgb}{1.0, 0.55, 0.41}

\def\lucas{\textcolor{black}}

\usepackage{todonotes}

\def\BibTeX{{\rm B\kern-.05em{\sc i\kern-.025em b}\kern-.08em
    T\kern-.1667em\lower.7ex\hbox{E}\kern-.125emX}}

\newcommand\copyrighttext{%
  \scriptsize © 2024 IEEE. The paper will appear in the IEEE Robotics and Automation Letters (RA-L). Personal use of this material is permitted.  Permission from IEEE must be obtained for all other uses, in any current or future media, including reprinting/republishing this material for advertising or promotional purposes, creating new collective works, for resale or redistribution to servers or lists, or reuse of any copyrighted component of this work in other works.
  }
\newcommand\copyrightnotice{%
\begin{tikzpicture}[remember picture,overlay]
\node[anchor=south,yshift=10pt] at (current page.south) {\fbox{\parbox{\dimexpr\textwidth-\fboxsep-\fboxrule\relax}{\copyrighttext}}};
\end{tikzpicture}%
}

\begin{document}

\title{Under-Canopy Navigation using Aerial Lidar Maps}

\author{Lucas Carvalho de Lima$^{1,2}$, Nicholas Lawrance$^{2}$, Kasra Khosoussi$^{2}$, Paulo Borges$^{2}$, Michael Br{\"u}nig$^{1}$%
\thanks{Manuscript received: February, 5, 2024; Revised: May, 6, 2024; Accepted: June, 5, 2024. This paper was recommended for publication by Editor Giuseppe Loianno upon evaluation of the Associate Editor and Reviewers' comments.}
\thanks{$^{1}$ L. de Lima and M. Br\"{u}nig are with the School of Electrical Engineering and Computer Science, The University of Queensland, Brisbane, QLD 4072, Australia}
\thanks{$^{2}$ L. de Lima, N. Lawrance, K. Khosoussi and P. Borges are with CSIRO Robotics, Data61, Pullenvale Qld 4069 Australia. All correspondence should be addressed to {\tt\footnotesize lucas.lima@data61.csiro.au}
}
\thanks{Digital Object Identifier (DOI): see top of this page.}
}

\markboth{IEEE Robotics and Automation Letters. Preprint Version. Accepted June, 2024}
{Lima \MakeLowercase{\textit{et al.}}: Under-Canopy Navigation using Aerial Lidar Maps} 

\maketitle

\copyrightnotice

\begin{abstract}

Autonomous navigation in unstructured natural environments poses a significant challenge. In goal navigation tasks without prior information, the limited look-ahead of onboard sensors utilised by robots compromises path efficiency. We propose a novel approach that leverages an above-the-canopy aerial map for improved ground robot navigation. Our system utilises aerial lidar scans to create a 3D probabilistic occupancy map, uniquely incorporating the uncertainty in the aerial vehicle's trajectory for improved accuracy. Novel path planning cost functions are introduced, combining path length with obstruction risk estimated from the probabilistic map.
The D* Lite algorithm then calculates an optimal (minimum-cost) path to the goal. This system also allows for dynamic replanning upon encountering unforeseen obstacles on the ground. Extensive experiments and ablation studies in simulated and real forests demonstrate the effectiveness of our system.
\end{abstract}

\section{Introduction}

\begin{figure}[t]
 \centering
  \includegraphics[width=.48\textwidth]{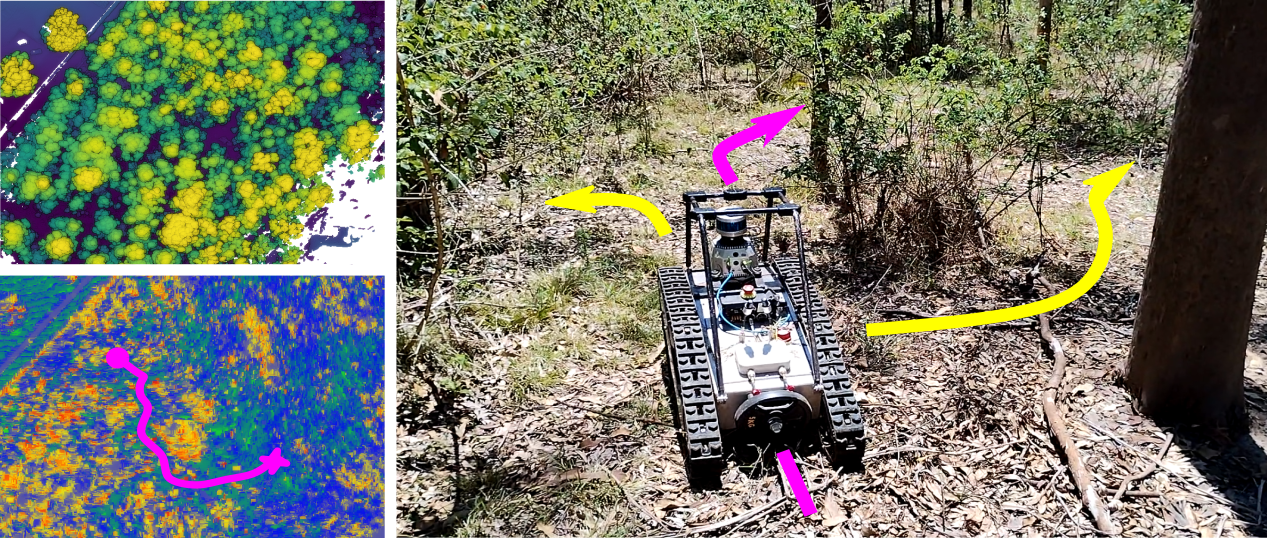}%
  \caption{\small{Sensors mounted on a ground robot can suffer from limited range and occlusions. This local view often precludes efficient long-distance navigation in complex scenarios, leading to dead-ends or unsafe paths (right figure: yellow arrows). Aerial lidar data (top left figure) acquired prior to operation can provide meaningful information for global planning (bottom left and right figures: magenta paths) allowing efficient navigation.}}
\label{fig:front_figure_v2}
 \end{figure}

\IEEEPARstart{L}{ong}-range navigation in natural environments such as forests remains a hard problem. Despite recent advances in perception and planning, ground robots may struggle to identify traversable areas in vegetated environments and the onboard sensor's limited field of view can lead to inefficient paths, frequently facing dead-ends during deployment. Aerial data, \eg{} lidar or satellite images, collected prior to the ground vehicle deployment, could offer a more global perspective, potentially allowing the generation of paths that avoid some of the obstacles visible from above (\figref{fig:front_figure_v2}). 

Although the problem of ground navigation using aerial information has been investigated previously \cite{Miller2022, Sharma2020}, leveraging airborne data for under-canopy robotic navigation poses several challenges. The different perspectives between above- and under-canopy views can lead to the occlusion of ground obstacles from aerial images or under-sampling of ground lidar points (the relevant points for ground-vehicle planning) caused by canopy cover. Moreover, the sensing noise and inevitable error in the estimated aerial vehicle's trajectory further hinder the creation of accurate aerial maps that can guide the ground robot. Additionally, the path planner must incorporate occupancy belief encoded in the probabilistic map in addition to path length to minimise the risk of lengthy detours and, ultimately, mission failure.  

We address these problems in a novel framework for under-canopy navigation in forests assisted by above-canopy aerial lidar scans.
Our system builds a 3D probabilistic occupancy map from overhead lidar observations and uses occupancy probabilities to estimate the likelihood of ground-level obstruction throughout the map.
Standard occupancy mapping techniques approach the problem as a pure mapping task with perfect sensor poses. However, ignoring the inevitable error in aerial sensor's pose estimates (produced by, e.g., pose-graph optimisation) can result in poor mapping accuracy, especially at longer ranges in aerial scenarios (lidar hit-point position error increases approximately linearly with the product of angle and range). To address this issue, our mapping system explicitly takes into account pose-estimation uncertainty and averages occupancy probabilities over the set of possible aerial vehicle trajectories using Monte Carlo sampling.
 The aerial map is then used to plan a safe and efficient path for the ground robot before its deployment. We introduce two path planning cost functions designed to strike a balance between minimising the travel distance and avoiding potential obstructions (inferred from the occupancy map).
 Our path planner is based on D* Lite~\cite{koenig2005fast} which supports efficient replanning in case the ground robot encounters a previously unseen obstacle along its current path.
Finally, we evaluate the performance of our full system and its components in extensive experiments conducted in simulated and real complex forest environments.

In summary, the main contributions of this work are: \lucas{1) A method for navigation in dense forest environments using aerial lidar observations as prior information; 2) A mapping algorithm to estimate the obstructed areas under the canopy using airborne lidar data; 3) A Monte Carlo sampling method to propagate uncertainty from the aerial sensor trajectory to 3D occupancy maps; 4) Experimental evaluation in simulation and real-world experiments in a challenging forest area using a mobile tracked robot.}

\section{Related Work}
\lucas{\textit{A. Aerial mapping for ground navigation} - Prior work has proposed a range of approaches for generating traversability or cost maps for ground navigation from aerial data. Applied to forested areas,} Vandapel et al.~\cite{Vandapel2006} estimate costs from airborne lidar based on the ``vegetationess" metric (ratio of ground points and total number of points falling in a 2D grid region) and a vehicle mobility analysis. \lucas{Combinations of} geometric features (\eg{} ground elevation) and semantic terrain classes from airborne data (lidar, colour and multispectral images) \lucas{are used to estimate traversability or binary costs in~\cite{Silver2006,Miller2022}}. Similarly, in~\cite{Eder2023} semantic terrain classes, geometric features (\eg{} slope) and proprioceptive locomotion data (\eg{} velocities, IMU) of different robotic platforms are processed to generate traversal costs conditioned on the platform type and terrain class. Lastly, Stentz et al.~\cite{Sofman2006} learn high-cost features \lucas{from satellite images using ground-view lidar features for self-supervision.} 
Despite their relevance, \lucas{these works} neglect free-space information\cite{Vandapel2006,Silver2006,Sofman2006} and measurement uncertainty\cite{Miller2022,Vandapel2006,Silver2006} in their costmap definitions, which are crucial elements for traversability prediction. Our approach relies on explicit probabilistic reasoning over voxel occupancy (free, occupied, or \cblue{uncertain}), while also accounting for sensor noise and pose-estimation error. 
\lucas{We also incorporate probabilistic obstruction and path length in our proposed cost functions for planning with uncertain maps, unlike previous methods.}

\lucas{\textit{B. Sensor pose uncertainty in occupancy maps - }}Standard occupancy mapping approaches \cite{octomap} and many of its variants \cite{agha2019,Meadhra2019,wang2016fast} all assume that sensor poses are perfectly known at the time lidar measurements are acquired. However, this approximation can easily lead to high occupancy errors if localisation uncertainty increases.
To address this problem, it is possible to incorporate sensor and localisation uncertainty in Gaussian process (GP) occupancy mapping by considering inputs as noisy observations drawn from a Gaussian distribution and integrating them in the kernel function~\cite{Callaghan2010}. However, such modifications increase the computational cost of the approach by an order of magnitude, which may limit its application in large datasets even in offline settings. 

We revisit and extend the idea of using Monte Carlo (MC) sampling to incorporate localisation uncertainty in 2D occupancy maps.
This approach was first proposed in \cite{Blanco2008} and used more recently in \cite{joubert2015pose}.
The former samples trajectories from a particle filter to marginalise poses from occupancy distributions, whereas the latter performs MC integration in the log-odds occupancy ratio.
While we also marginalise out poses from occupancy probabilities using MC sampling, unlike \cite{Blanco2008,joubert2015pose} our approach generates samples from the approximated posterior of a modern pose-graph SLAM system. Furthermore,
we apply this idea to 3D mapping and conduct extensive experimental evaluations in simulated forest environments.   
 
\lucas{\textit{C. Path planning} - }Graph-based (D*, A*) or sampling-based (Rapidly Exploring Random Trees - RRT*) methods have been broadly used for global planning in costmaps \cite{Krusi2017}. When applied to occupancy maps, occupancy probabilities are often thresholded to classify grid cells as obstacle or free space, subsequently assigning high and low costs to them \cite{David2015, meyer2012}. Other works \cite{Vandapel2006, Timothy2020} propose heuristic functions of terrain features to build costmaps. However, planning on thresholded occupancy maps or feature-based costs without uncertainty information can create a false sense of confidence in the generated path, \lucas{potentially} leading to long detours. \lucas{Some authors address planning under uncertainty by minimising a risk metric over map predictions~\cite{STEP}. In contrast, Banfi et. al~\cite{Banfi2022} generate multiple path hypotheses in unknown areas, further reasoning on locally acquired information from a ``next-best view" (NBV) pose to take the best route. 
}

Unlike the above approaches, our system generates a minimum-cost path to the goal by minimising cost functions that blend obstruction risk with path length. Additionally, we use D* Lite to efficiently and optimally replan when the ground robot encounters previously unmapped obstacles along its current path.

\section{System Overview}

We divide the problem of forest navigation using uncertain prior maps from aerial lidar into two tasks: 1) offline mapping and global planning using airborne sensing data; and 2) online execution of waypoints and obstacle avoidance using onboard perception on a ground robot. Since this work is focused on offline mapping and planning, we utilise an off-the-shelf onboard navigation system (local navigation module~\cite{Hudson_2022}) for the online execution (task 2) in our experiments. 

 \begin{figure}[t]
 \centering
  \includegraphics[width=.48\textwidth]{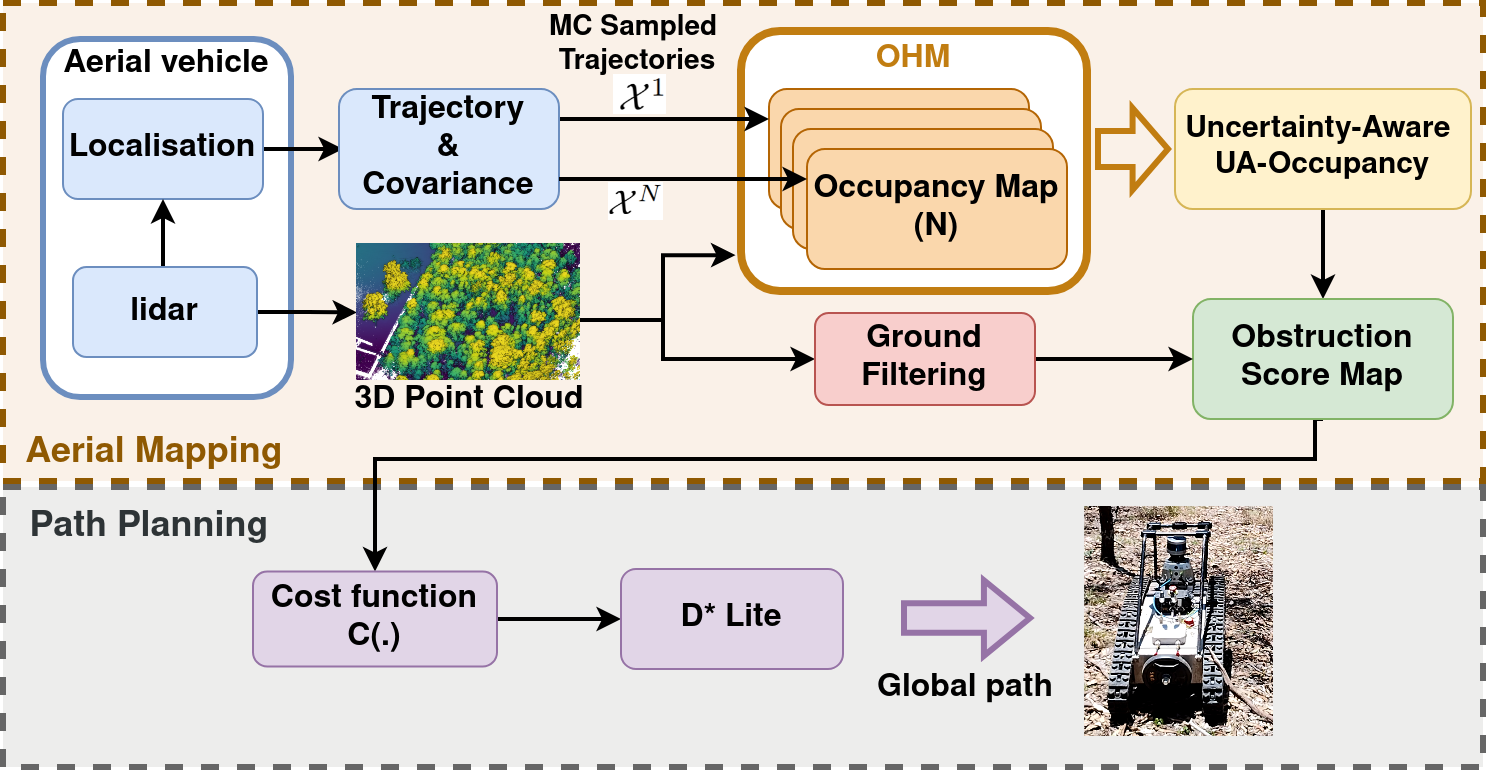}%
\caption{\small{Overview of our proposed navigation system. First, aerial lidar data are fused with uncertain sensor poses to generate a 3D probabilistic occupancy map. Next, the 3D map is further processed to generate a 2D ground obstruction map. Finally, obstruction scores are passed through a cost function to the D* Lite planner to produce global paths that guide the ground robot during operation.}}
\label{fig:diagram}
 \end{figure}

\figref{fig:diagram} summarises our system. The aerial mapping (\secref{sec:mapping}) aggregates airborne lidar scans and estimates the sensor trajectory using a localisation system (Wildcat SLAM \cite{ramezani2022wildcat}). 
Then, we estimate the ground height from the point cloud (\secref{sec:ground_filter}), in addition to building a 3D probabilistic occupancy map from the aerial lidar observations (\secref{sec:expected_occ_map}). For improved map accuracy, we also encode the uncertainty from sensor poses into occupancy distributions employing a Monte Carlo integration method. Next, the ground height estimates and occupancy values are fed into our scoring function to estimate the obstructed (non-traversable) areas over the terrain (\secref{sec:blocked_probability}).
Obstruction scores are finally integrated in two evaluated cost functions, expected cost and log-reachability cost, ultimately used by D* Lite to generate optimal paths.

\section{Aerial Mapping}
\label{sec:mapping}

\subsection{Ground Filtering}
\label{sec:ground_filter}

The ground filtering module takes in a 3D point cloud and produces a gridded ($0.25$ m side-length squares) 2D height map representing the terrain's supporting surface.
Height estimates are used in our mapping approach to estimate obstructed regions above the terrain. 

We first apply the Cloth Simulation Filtering (CSF)~\cite{Zhang2016} to segment ground points from the global point cloud. CSF vertically ($Z$) inverts the point cloud data and simulates gravity acting on a semi-rigid cloth -- modeled as interconnected mass-spring nodes -- covering the inverted surface. Then, CSF segments ground points by retaining the closest points in the cloud to the CSF nodes within a height threshold ($h_{cc}$) in each grid cell. For cells without ground points, we use the closest CSF node as a virtual ground. Finally, the minimum height point in each grid cell is used as the ground height estimate. By operating on inverted point clouds, CSF is less susceptible to segmentation errors (\eg{} lidar canopy reflections or high vegetation mistakenly classified as ground) and generalises well to distinct environments using only few parameters, \lucas{\ie{} cloth rigidness ($RI = 3$) and height threshold ($h_{cc} = 0.20m$)}.

\subsection{Occupancy mapping with imperfect pose estimates}
\label{sec:expected_occ_map}
In addition to the ground elevations, occupancy estimates of the environment are also input to the obstruction estimation module. We process lidar points above ground using Occupancy Homogeneous Mapping (OHM) \cite{ohm} to build an occupancy grid map. OHM estimates occupancy probability by taking into account lidar hits and pass-throughs for each 3D grid element (voxel). Standard occupancy mapping methods such as OHM assume lidar poses are perfectly known. However, in practice, poses are estimated by other state-estimation systems (\eg~\lucas{SLAM}, pose-graph optimisation, \lucas{GNSS}) which are inevitably subject to estimation errors due to sensor noise. Ignoring pose estimation error leads to  inaccurate maps. We address this challenge by incorporating the estimation uncertainty of the aerial vehicle's trajectory from the pose-estimation system into occupancy mapping.

The occupancy mapping module estimates the occupancy probability of each voxel given aerial lidar measurements. We denote the map by $\{p(m_i \mid \mathcal{Z})\}_i$ where $m_i$ is a Bernoulli random variable indicating the occupancy of the $i$th voxel, and $\mathcal{Z}$ denotes the set of \emph{all} measurements from the aerial vehicle (e.g. lidar, IMU, and GNSS). Voxel occupancy probability is
\begin{align}
   p(m_i \mid \mathcal{Z}) = \int p(m_i \mid \mathcal{X}, \mathcal{Z}) \, p(\mathcal{X} \mid \mathcal{Z}) \, \mathrm{d}\mathcal{X},    
   \label{eq:occupancy_integral}
\end{align}
where $\mathcal{X}$ denotes the robot's trajectory (represented by a finite set of poses). 
The first integrand $p(m_i \,|\, \mathcal{X}, \mathcal{Z})$ in \eqref{eq:occupancy_integral} is the voxel occupancy probability computed by standard occupancy mapping systems given trajectory $\mathcal{X}$. 
Note that voxel occupancy $m_i$ is conditionally independent from sensor measurements given the trajectory $\mathcal{X}$. This term can be further simplified to $p(m_i \,|\, \mathcal{X}, \mathcal{Z}_\text{lidar})$ where $\mathcal{Z}_\text{lidar} \subset \mathcal{Z}$ denotes the set of lidar measurements. The second integrand $p(\mathcal{X} \,|\, \mathcal{Z})$ is the posterior distribution over the trajectory given all measurements. We use Monte Carlo (MC) integration to approximate the intractable integral in~\eqref{eq:occupancy_integral} by sampling $N$ \lucas{independent} trajectories $\{\mathcal{X}^{[k]}\}_{k=1}^{N}$ from the \lucas{trajectory posterior, where $\mathcal{X}^{[k]} \sim p(\mathcal{X} \mid \mathcal{Z})$,} and computing the following expression,
\begin{equation}
  p(m_i \mid \mathcal{Z}) \approx  \frac{1}{N} \sum_{k=1}^N p(m_i \mid \mathcal{X}^{[k]}, \mathcal{Z}_\text{lidar}).
   \label{eq:occ_sampled}
\end{equation}
Equation \eqref{eq:occ_sampled} suggests the following scheme for incorporating pose-estimation uncertainty into occupancy mapping: (i) build an occupancy map using a standard occupancy mapping method such as OHM for each trajectory sample $\mathcal{X}^{[k]}$; and (ii) for each voxel compute the average of the 
resulting $N$ predicted occupancy probabilities. 
We term this approach \lucas{\emph{Uncertainty-Aware (UA) occupancy mapping}}.

\lucas{Note that the details of trajectory sampling depend in part on the underlying pose-estimation technique. For example, one may use ancestral sampling to sample a trajectory when the pose-estimation is obtained by integrating inertial measurements over time. If poses are estimated independently using absolute measurements (e.g., GPS), we just need to independently sample  poses from the corresponding posteriors to obtain a trajectory sample. More complex pose-estimation systems such as SLAM typically provide a Gaussian approximation of pose marginals from which one can sample poses independently to obtain an approximate trajectory sample. Alternatively, one can avoid the error caused by breaking the correlated trajectory posterior into pose marginals by \emph{transporting} samples generated from the standard normal distribution using the inverse Cholesky factor of the sparse information matrix provided by SLAM systems.}

\subsection{Obstruction score estimation}
\label{sec:blocked_probability}
 \begin{figure}[t]
 \centering
  \includegraphics[width=.48\textwidth]{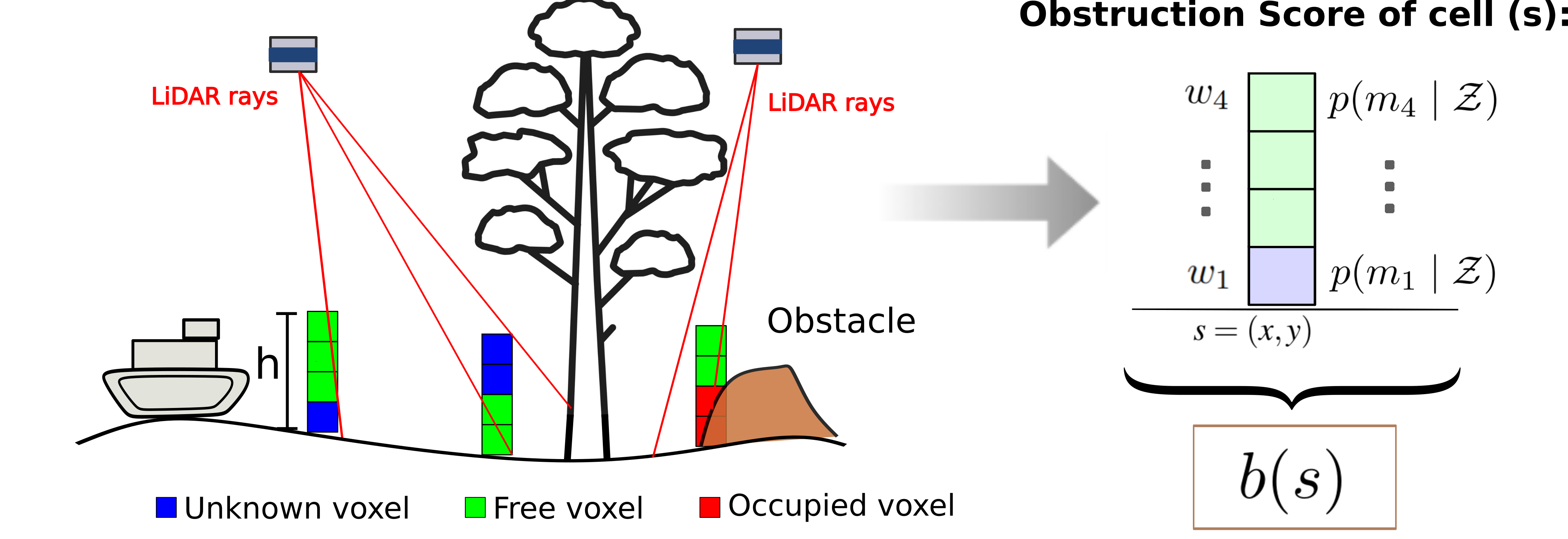}%
\caption{\small{We reason on free, occupied and \cblue{uncertain} voxels by integrating lidar measurements in an occupancy grid map. Later the occupancy probabilities are fused in our scoring method to estimate the final obstruction map.}}
\label{fig:mapping_example}
 \end{figure}

We use our 3D \lucas{\ourmap} map and the estimated ground height to assign a simple obstruction risk score to each ground-level 2D cell. This score is computed for each vertical column of $n$ 3D voxels starting from the ground cell as illustrated in \figref{fig:mapping_example}. 
Formally, the score function $b: \mathbb{Z} \times \mathbb{Z} \to [0,1]$ assigns a value between zero and one to each 2D cell (i.e., the grid $XY$ coordinates of the corresponding column in the world frame). A larger value of $b(s)$ for a cell $s \triangleq (x,y)$ indicates a higher risk of obstruction at that cell.
The number $n$ of column voxels is chosen according to the clearance height ($h$ in \figref{fig:mapping_example}) required by the robot to pass through a column. Hence, taller robots require a larger $n$.

A cell $s$ might be considered as obstructed if at least one of the voxels in the corresponding column is occupied. This choice leads to $b(s) = 1-\prod_{i \in \mathcal{I}_s} p(m_i = 0 \mid \zcal)$. However, because of the complexity of forest environments and the sparsity of aerial lidar rays reaching the ground level, the resulting score function will likely fail to correctly capture traversable cells where only the first voxel above ground is occupied by grass or an overhanging foliage is occupying the top voxel in the stack. An effective heuristic estimate is to instead use the weighted average of voxel occupancy probabilities within a column as the corresponding obstruction score, where the (fixed) weights reflect the significance of voxel occupancy. The resulting score function $b$ is given by
\begin{equation} 
 b(s) \triangleq
\sum_{i \in \mathcal{I}_s} \frac{w_i}{\sum_{i \in \mathcal{I}_s} w_i} \, p(m_i = 1 \mid \mathcal{Z}),
 \label{eq:trav_prob}
\end{equation}
where $\mathcal{I}_s$ denotes the set of indices of the $n$ voxels that are part of the vertical stack at cell $s$, and $p(m_i = 1 \mid \mathcal{Z})$ is the \lucas{\ourmap} probability of voxel $i$ given the aerial measurements. Specifically, in our work $n = 4$ which allows for a robot height of $1\,$m using a $0.25\,$m voxel size, and the voxels weights $w_i$ from bottom to top are set to $(1, 2, 2, 2)$. The rationale behind these values is to assign a smaller weight to the first voxel above ground level since the tracked mobile robot used in our experiments can better navigate obstacles immediately above the ground (\eg{} small rocks, grass). %

Finally, note that the obstruction score $b(s)$ is computed for the corresponding voxel-sized 2D cell $s$ which is typically much smaller than the robot's footprint. Therefore, in the definition of path costs in \secref{sec:planning}, we use the (worst-case) footprint obstruction score $b_{\text{max}}(s) \triangleq \max_{s_i \in \fcal(s)} \, b(s_{i})$ among cells in the robot footprint $\mathcal{F}(s)$ when the robot's center is inside cell $s$.
For convenience, we use a circular footprint, avoiding the need to consider different heading angles of the robot. 
\section{Global Path Planning}
\label{sec:planning}
Our path planning module plans a 2D path (over the ground cells) for the ground robot given a pair of start and goal locations. A path $\mathcal{P}$ is a sequence of adjacent ground cells $\mathcal{P} = (s_0,s_1,\ldots,s_{\ell})$ starting from the current cell $s_0$ and ending at the goal cell $s_\ell = s_\text{goal}$.
The cost of transitioning from cell $s$ to one of its adjacent cells $s'$ (hereafter, the edge cost) is denoted by $c(s,s')\in \mathbb{R}^+$. We use D* Lite \cite{koenig2005fast} to find a minimum-cost path where the path cost is given by the sum of its edge costs 
$J({\mathcal{P}}) \triangleq \sum_{i=0}^{\ell-1} c(s_i, s_{i+1})$.
D* Lite allows us to efficiently replan an optimal path from the current cell to the goal based on the latest edge costs upon encountering a blockage.
In the following, we introduce two choices used in our simulations and experiments for the edge costs $c(s,s')$ that incorporate both the footprint obstruction score $b_\text{max}(s)$ (reflecting the risk of choosing a blocked path) and the travelled distance.
\subsubsection{Expected Cost}
This edge cost function is inspired by the expected traversal cost proposed in \cite[Eq.~8]{Missiuro06}. Formally, the edge cost between two adjacent cells $s_i$ and $s_{i+1}$ in a given path is defined as follows,
\begin{equation} 
c(s_i, s_{i+1}) = b_{\text{max}}(s_{i+1}) \, C_\text{obst} + \left(1 - b_{\text{max}}(s_{i+1})\right) ||s_{i+1} - s_{i}||
\label{eq:expected_cost}
\end{equation}
where $C_\text{obst} \in \mathbb{R}^+$ is a penalty parameter aimed at discouraging the robot from entering cells that have a high obstruction score. Higher values of $C_\text{obst}$ result in a higher penalty for overlapping with potentially obstructed cells. Intuitively, if we interpret $b_\text{max}(s) \in [0,1]$ as the obstruction probability when the robot's center is inside cell $s$, the above expression becomes the expected value of the following function:
\begin{equation}
    (s_i,s_{i+1}) \mapsto \begin{cases}
    C_\text{obst} & \text{if $s_{i+1}$ is blocked,}\\
    \|s_{i+1} - s_i\| & \text{if $s_{i+1}$ is free.}
    \end{cases}
\end{equation}
In this way, the cost function represents an expectation of the transition cost where $C_\text{obst}$ is the additional detour cost incurred when overlapping an obstructed cell, and $b_\text{max}(s)$ is the likelihood. 
D* Lite also requires a heuristic cost %
function $h(s_0, s_i)$ that must be non-negative and backward consistent, \ie{} it obeys $h(s_0, s_0) = 0$ and $h(s_0, s_i) \leq h(s_0, s_{i-1}) + c(s_{i-1},s_i)$. In our method, the heuristic cost is defined as
\begin{equation} 
h(s_0, s_i) = (1 - b_{\text{min}})d(s_0,s_{i}),
\label{eq:expected_heuristic}
\end{equation}
where $d(s,s')$ is the octile\footnote{Octile distance: $d(s, s') \triangleq \max(\Delta x,\Delta y) + (\sqrt{2} - 1)\min(\Delta x,\Delta y)$, where $\Delta x = |s'[0] - s[0]|$ and $\Delta y = |s'[1] - s[1]|$.} distance between $s$ and $s'$ considering also diagonal movements in a grid, and $b_{\text{min}} \triangleq \min_s \, b(s)$ is the minimum obstruction score found in our map.

\subsubsection{Log-Reachability cost}
This path cost is based on the reachability metric proposed in \cite{Heiden17}. Informally speaking, reachability of a path reflects the likelihood of reaching the goal cell without hitting any blocked cells.
We approximate the product integral derived in \cite[Eq.~14]{Heiden17} for computing the reachability of the goal path $\mathcal{P}$ by the following expression,
\begin{equation} 
R(\mathcal{P}) \approx \prod_{i = 0}^{\ell-1}\left(1 - b_{\text{max}}({s_{i+1}})\right)^{||s_{i+1} - s_{i}||}.   
\label{eq:reach_final}
\end{equation}
We find the optimal path by minimising the negative log-reachability, yielding the following edge cost, 
\begin{align}
c(s_i,s_{i+1}) &= -\log\left(1 - b_{\text{max}}(s_{i+1})\right) \,{||s_{i+1} - s_{i}||}.
\label{eq:reachability_cost}
\end{align}
The heuristic cost function used for the reachability cost is
\begin{align}
    h(s_0, s_i) = -\log(1 - b_{\text{min}}) \, d(s_0,s_i)
    \label{eq:log_heuristic}
\end{align}
where, as before, $d(s,s')$ is the octile distance between $s$ and $s'$, and $b_\text{min}$ is the minimum obstruction score in the map.

\section{Simulations}
We validated our proposed approach using the Gazebo simulator\footnote{\url{https://gazebosim.org/}}. We created synthetic forests using vegetation models of the NEGS-UGV Dataset~\cite{sanchezNEGS2022}, and acquired simulated aerial lidar scans of these forest worlds. In the following, we first evaluate the accuracy of our proposed \lucas{UA-}occupancy mapping, and then analyse the performance of our full navigation system. 

\subsection{Ablation Study: UA-Occupancy Mapping}
\label{sec:map_results}
\begin{figure}[t]
 \begin{center}
  \subfloat[]{\includegraphics[width=.24\columnwidth]{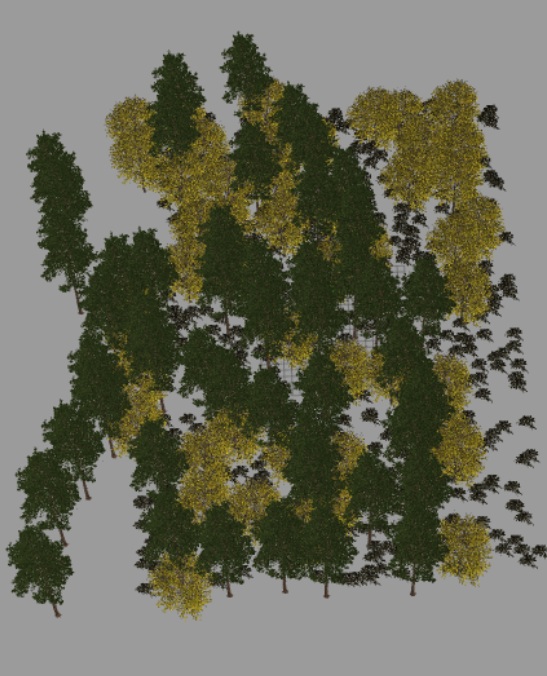}%
  } \hspace{0.01mm}
  \subfloat[]{\includegraphics[width=.24\columnwidth]{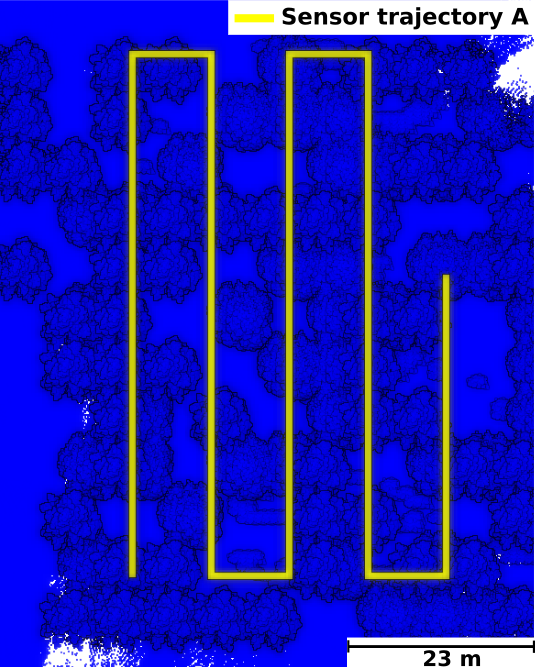}%
  }\hspace{0.01mm}
    \subfloat[]{\includegraphics[width=.24\columnwidth]{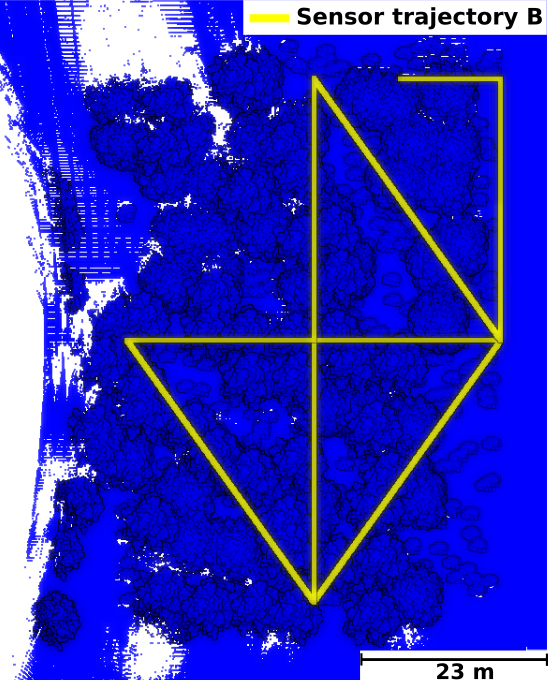}%
  } \hspace{0.01mm}
  \subfloat[]{\includegraphics[width=.24\columnwidth]{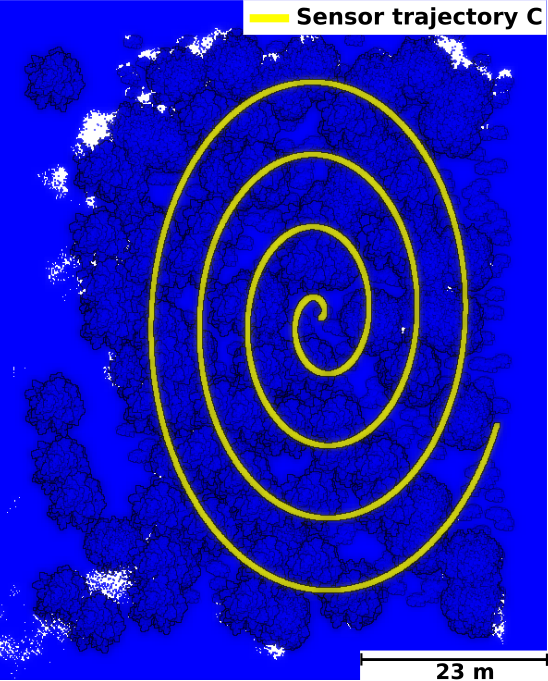}%
  }
\end{center}
 \caption{\small{(a): Example of synthetic forest created in Gazebo \lucas{where we simulated lidar scans captured above the canopy.} (b) to (d): Ground-truth \lucas{trajectories (A, B and C)} of the airborne sensor and the corresponding registered scans for the datasets acquired in the synthetic environments \forestA{}, \textit{II} and \textit{III}, respectively. After adding perturbations to sensor poses, a perturbed point cloud of each dataset was produced to be processed by our approach.}}
 \label{fig:gazebo_forests}
 \end{figure}

First, we evaluate the accuracy of our \lucas{UA-occupancy} map (with \lucas{pose} uncertainty) and compare it with the standard occupancy map for different levels of error in the sensor trajectory.
We created three random forest worlds: \forestA, \textit{II} and \textit{III}, and simulated three different aerial trajectories \lucas{(one per scenario)}, as shown in \figref{fig:gazebo_forests}, of a VLP-16 lidar sensor flying above the canopy and scanning the forests. \lucas{We also simulated trajectory A in \textit{Forest II} and \textit{III}}. We constructed ground-truth occupancy maps for each scenario by simulating an ideal sensor (no noise in range measurements or trajectory) and processing the ground-truth point clouds and trajectories using OHM. Then, we perturbed each sensor pose along the trajectories with independent, additive Gaussian noise $\sim \mathcal{N}\left(0, \sigma^2\right)$ on each component \lucas{$[x,y,z,\text{roll}, \text{pitch}, \text{yaw}]$} to simulate a \lucas{GPS-like} \lucas{estimate} of the sensor trajectory. \lucas{Therefore, perturbations are not accumulated along the trajectory.} For each perturbation level, specified by standard deviation values, we generated 10 perturbed trajectories and point clouds for subsequent comparisons. We added Gaussian noise with a fixed standard deviation ($\sigma_r = 0.01\,$m) to range measurements to simulate sensor uncertainty. For our \lucas{\ourmap} mapping (\secref{sec:expected_occ_map}) we chose $N=20$. 
With full knowledge of ground-truth occupancy probabilities from the ideal sensor, and the perturbed datasets, we can further evaluate map accuracy using the Kullback-Leibler Divergence (KLD) of voxel occupancy \lucas{values} between \lucas{our UA} and standard occupancy maps with respect to the ground-truth.

In \forestA{}, we analysed the isolated effect of translation and orientation perturbations. Since the percentage of occupied voxels (about $10\%$) in the ground-truth map is considerably lower than the free counterparts (about $80\%$), we also computed the average KLD of \lucas{our UA} and standard occupancy maps per voxel class (free, occupied and \cblue{uncertain}) in this first scenario. Ground-truth probability ($p_\text{gt}$) thresholds are used to classify voxels as free ($p_\text{gt} < 0.45$), \cblue{uncertain} ($0.45 \leq p_\text{gt} \leq 0.55$) or occupied ($p_\text{gt} > 0.55$). 

\begin{figure}[!ht]
 \begin{center}
  \subfloat[]{\includegraphics[width=.49\columnwidth]{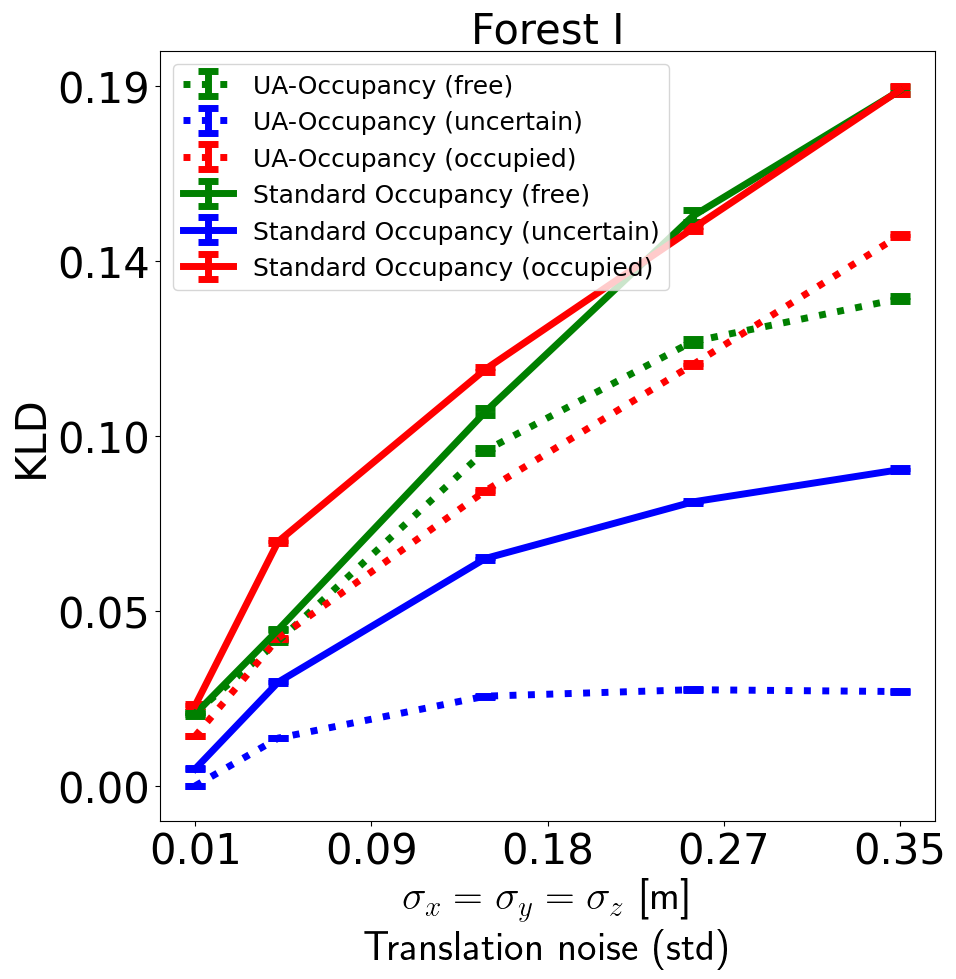}%
  } \hspace{0.1mm}
  \subfloat[]{\includegraphics[width=.49\columnwidth]{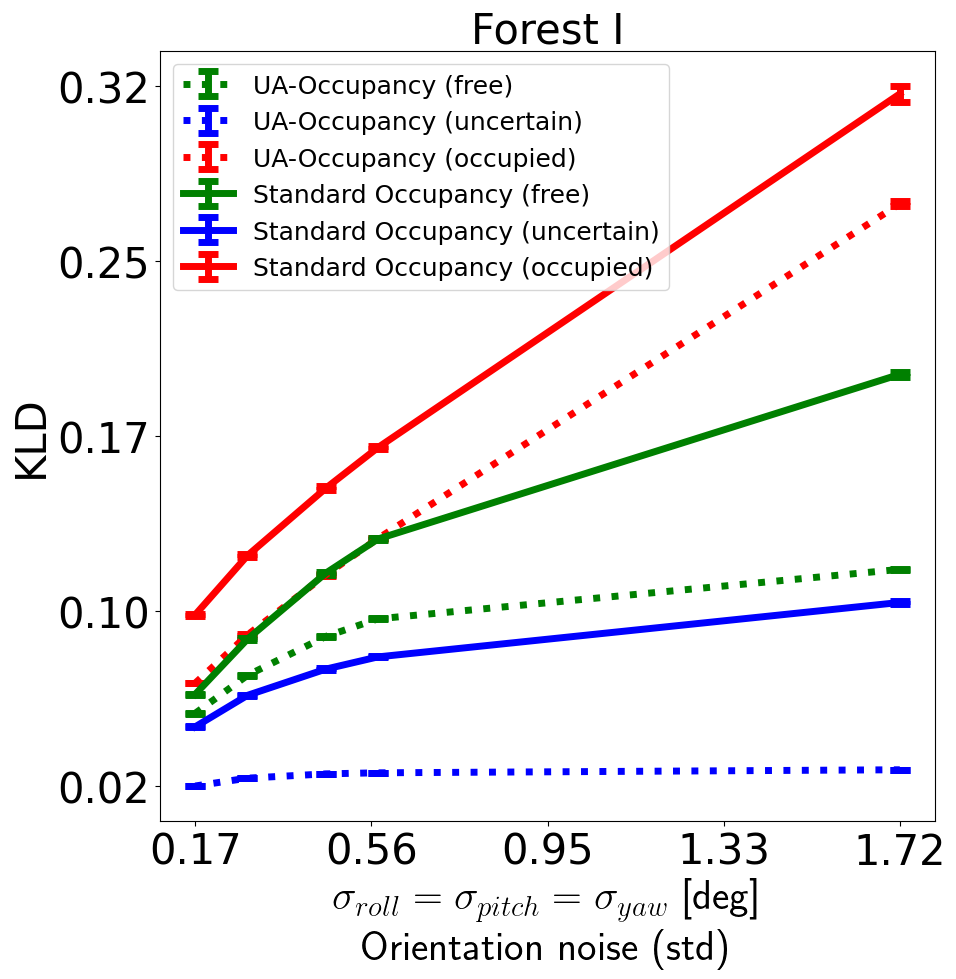}%
  }\\
    \subfloat[]{\includegraphics[width=.49\columnwidth]{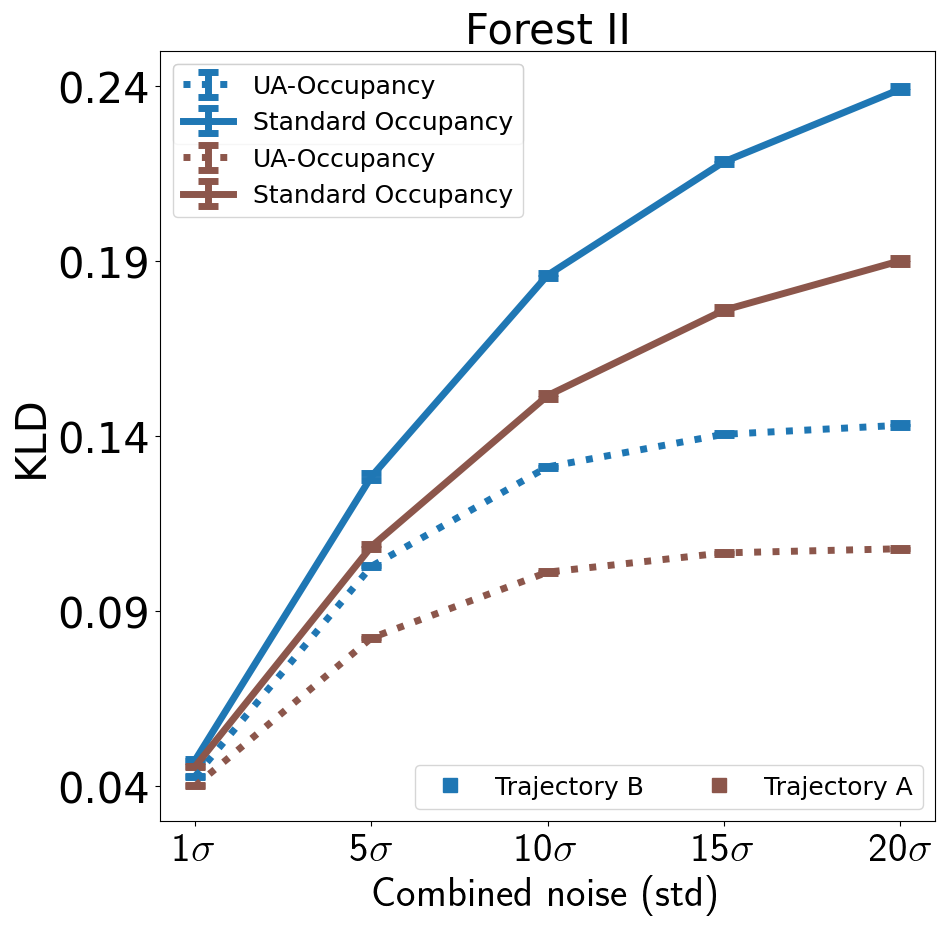}%
  } \hspace{0.1mm}
  \subfloat[]{\includegraphics[width=.49\columnwidth]{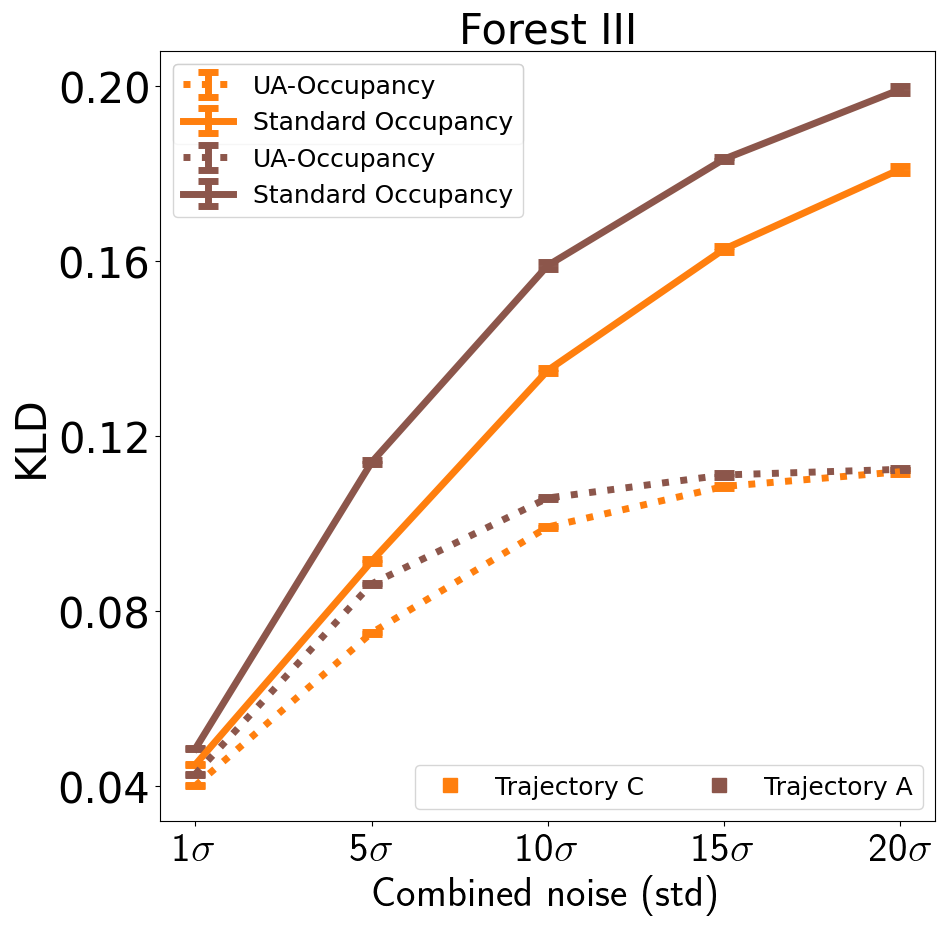}%
  }
\end{center}
 \caption{\small{Comparison of \lucas{our UA-occupancy} against the standard occupancy map for different perturbation values added to sensor poses. (a) and (b) show the average KLD values in \forestA{} \lucas{(trajectory A)} for perturbations in position only and orientation only, respectively. KLD values are depicted per voxel class on the ground-truth dataset. (c) and (d) illustrate the KLD values for all voxel classes in \forestB{} \lucas{(2 trajectories)} and \forestC{} \lucas{(2 trajectories)}, respectively, for combined perturbation levels (position and orientation) where $\sigma=(0.015 \text{m}, 0.015 \text{m}, 0.015 \text{m}, 0.11^{\circ} , 0.11^{\circ}, 0.11^{\circ})$. In the figures, noise refers to perturbations added to sensor poses and \lucas{error bars show $\pm$ std. deviation}.}}
 \label{fig:kld_plots}
 \end{figure}
Results illustrated in \figref{fig:kld_plots} (a) and (b) show that for both perturbations (position and orientation), the maps generated using \lucas{our \ourmap} presented lower average KLD for all perturbation levels, and that this behaviour was consistent among all voxel classes, confirming that our method improves the quality of maps on occupied, free and \cblue{uncertain} regions. We also assessed the effect of combined perturbations (position and orientation jointly) on map quality in \forestB{} and \textit{III} (\figref{fig:kld_plots} (c) and (d), respectively). As observed, the average KLD (all voxels) on the \lucas{\ourmap} also outperforms the standard occupancy with combined perturbation values.

Finally, we computed the Area Under Curve (AUC) of the Receiver Operating Characteristic (ROC) curves for the standard occupancy and \lucas{\ourmap} approaches, providing an overview of each method's performance as predictors of free or occupied states in datasets \textit{Forest II} and \textit{III}. Results are summarised in \tabref{table:AUC}, where AUC values closer to one indicate better prediction rates. For increasing levels of perturbation, both approaches experience reduction of AUC, which is anticipated since higher errors on sensor poses negatively affect the quality of occupancy values in both methods. However, in both datasets, the \lucas{\ourmap} method displayed higher AUC than the standard occupancy, showing the benefits of \lucas{fusing pose} uncertainty in the occupancy calculation.

\begin{table}[t]
\centering
\caption{Comparison of the AUC values ($\pm$ std. deviation) between the standard occupancy (standard) and UA-occupancy (ours) approaches for increasing perturbation levels in sensor poses, where $\sigma=(0.015 \text{m}, 0.015 \text{m}, 0.015 \text{m}, 0.11^{\circ} , 0.11^{\circ}, 0.11^{\circ})$. Higher AUC values indicate better map predictions.}
\resizebox{\columnwidth}{!}{%
{\color{black}\begin{tabular}{c l l l l} 
 \hline
  \multicolumn{1}{c}{Perturbation} & \multicolumn{2}{c}{\forestB{} - \textit{Trajectory B}} & \multicolumn{2}{c}{\forestC{} - \textit{Trajectory C}}\\
 \multicolumn{1}{c}{level} & \multicolumn{1}{c}{standard} & \multicolumn{1}{c}{ours} & \multicolumn{1}{c}{standard} & \multicolumn{1}{c}{ours} \\
 \hline
 $\sigma$   & 0.982 $\pm$ \num{09e-5} & \textbf{0.989} $\pm$ \num{04e-5} & 0.984 $\pm$ \num{11e-5} & \textbf{0.991} $\pm$ \num{06e-5} \\
 5$\sigma$  & 0.954 $\pm$ \num{48e-5} & \textbf{0.966} $\pm$ \num{17e-5} & 0.965 $\pm$ \num{35e-5} & \textbf{0.978} $\pm$ \num{9e-5}\\
 10$\sigma$ & 0.924 $\pm$ \num{52e-5} & \textbf{0.944} $\pm$ \num{18e-5} & 0.943 $\pm$ \num{30e-5} & \textbf{0.962} $\pm$ \num{11e-5} \\
 15$\sigma$ & 0.897 $\pm$ \num{59e-5} & \textbf{0.924} $\pm$ \num{15e-5} & 0.920 $\pm$ \num{70e-5} & \textbf{0.946} $\pm$ \num{29e-5} \\
 20$\sigma$ & 0.870 $\pm$ \num{87e-5} & \textbf{0.906} $\pm$ \num{45e-5} & 0.897 $\pm$ \num{67e-5} & \textbf{0.932} $\pm$ \num{35e-5} \\ 
  \hline
\end{tabular}
}}
\label{table:AUC}
\end{table}

\subsection{Full system evaluation}
To assess whether our navigation approach (mapping and planning), produces more efficient paths in comparison with a \lucas{na\"{\i}ve} baseline planner without prior maps, we first created a simulated forest with ground obstacles. \lucas{Then, we simulated two types of aerial sensor trajectory-estimation with perturbed poses, a GPS-only trajectory similar to \secref{sec:map_results} using $\sigma_{\textit{GPS}}=(0.05 \text{m}, 0.05\text{m}, 0.05\text{m}, 0.1^{\circ}, 0.1^{\circ}, 0.1^{\circ})$, and a SLAM-like trajectory where perturbations (added to relative pose transformations) are accumulated along the trajectory to simulate odometry drift. Absolute corrections (\eg{} perturbed GPS observations) are also added at each 120 sec to limit drift. The Gaussian noise added to relative transformations is defined by $\sigma_{\textit{drift}} = (0.009 \text{m}, 0.009\text{m}, 0.002\text{m}, 0.0057^{\circ}, 0.0057^{\circ}, 0.0057^{\circ})$ and the absolute observation noise defined by $\sigma_{\textit{abs}} = (0.05 \text{m}, 0.05\text{m}, 0.05\text{m}, 0.1^{\circ}, 0.1^{\circ}, 0.1^{\circ})$.} We then processed the lidar scans using our mapping method (\secref{sec:mapping}) to create the obstruction map used by the D* Lite planner. \lucas{For the MC integration (\secref{sec:expected_occ_map}), we sample poses independently from the GPS-like trajectory, whereas pose-sampling from the SLAM-like trajectory uses ancestral sampling (odometry observations) and independent sampling (absolute observations).}

\begin{figure}[ht]
\centering
  \subfloat[]{\includegraphics[width=.45\columnwidth, valign=t]{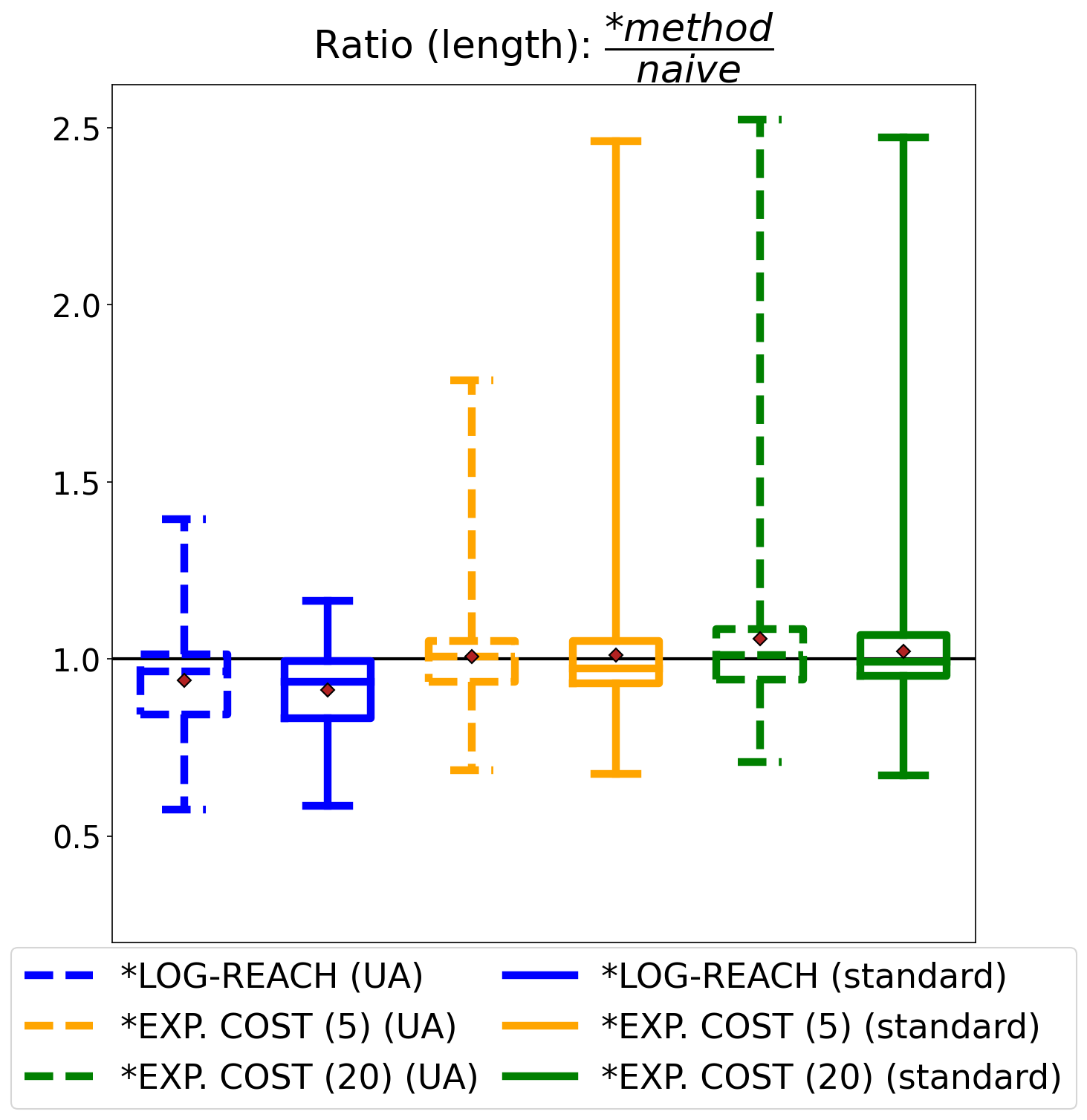}%
  }  \hspace{0.1mm}
    \subfloat[]{\includegraphics[width=.47\columnwidth, trim={0.5cm 0 0 1.1cm},clip, valign=t]{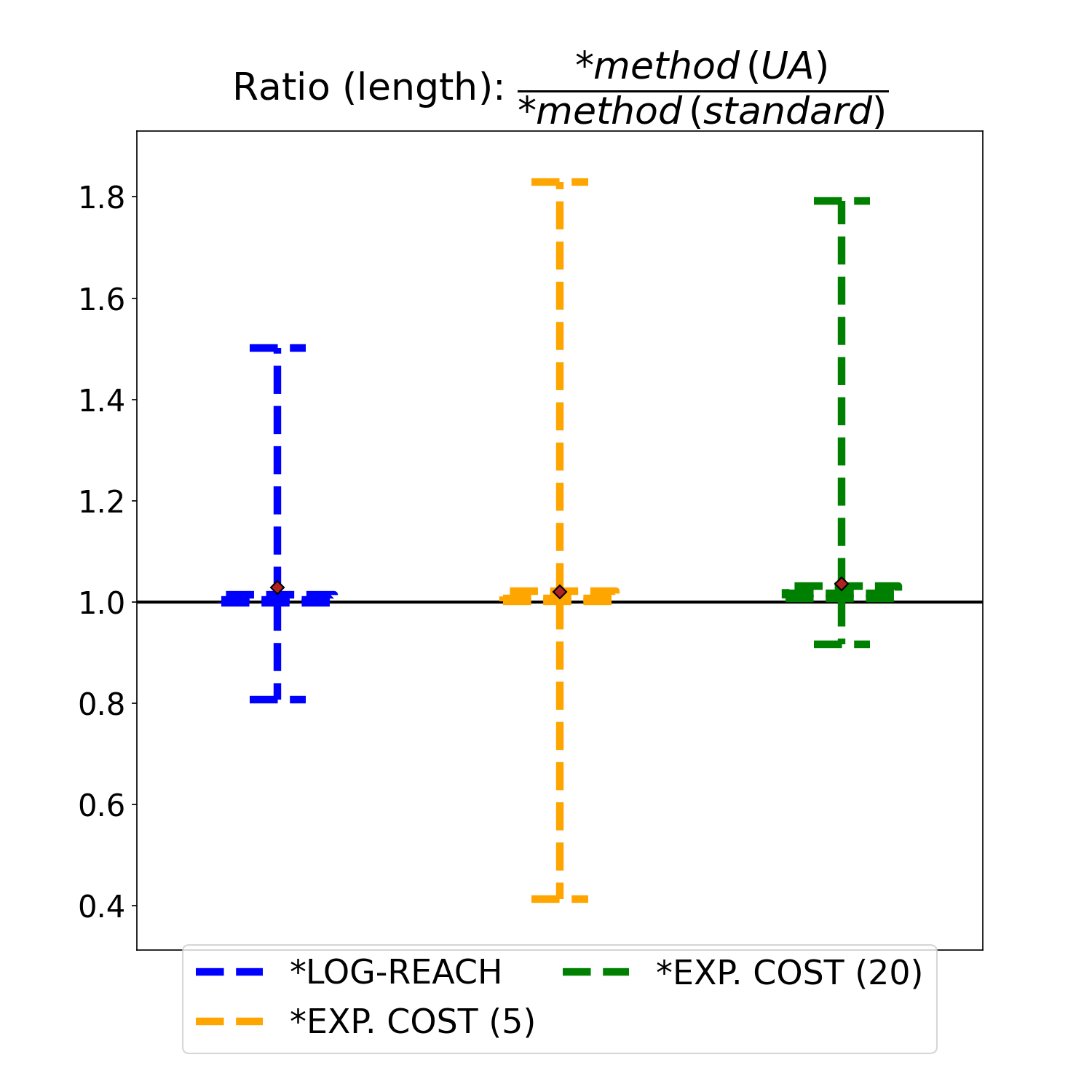}%
    } \vspace{1mm}
    
    \subfloat[]{\includegraphics[width=.45\columnwidth, valign=t]{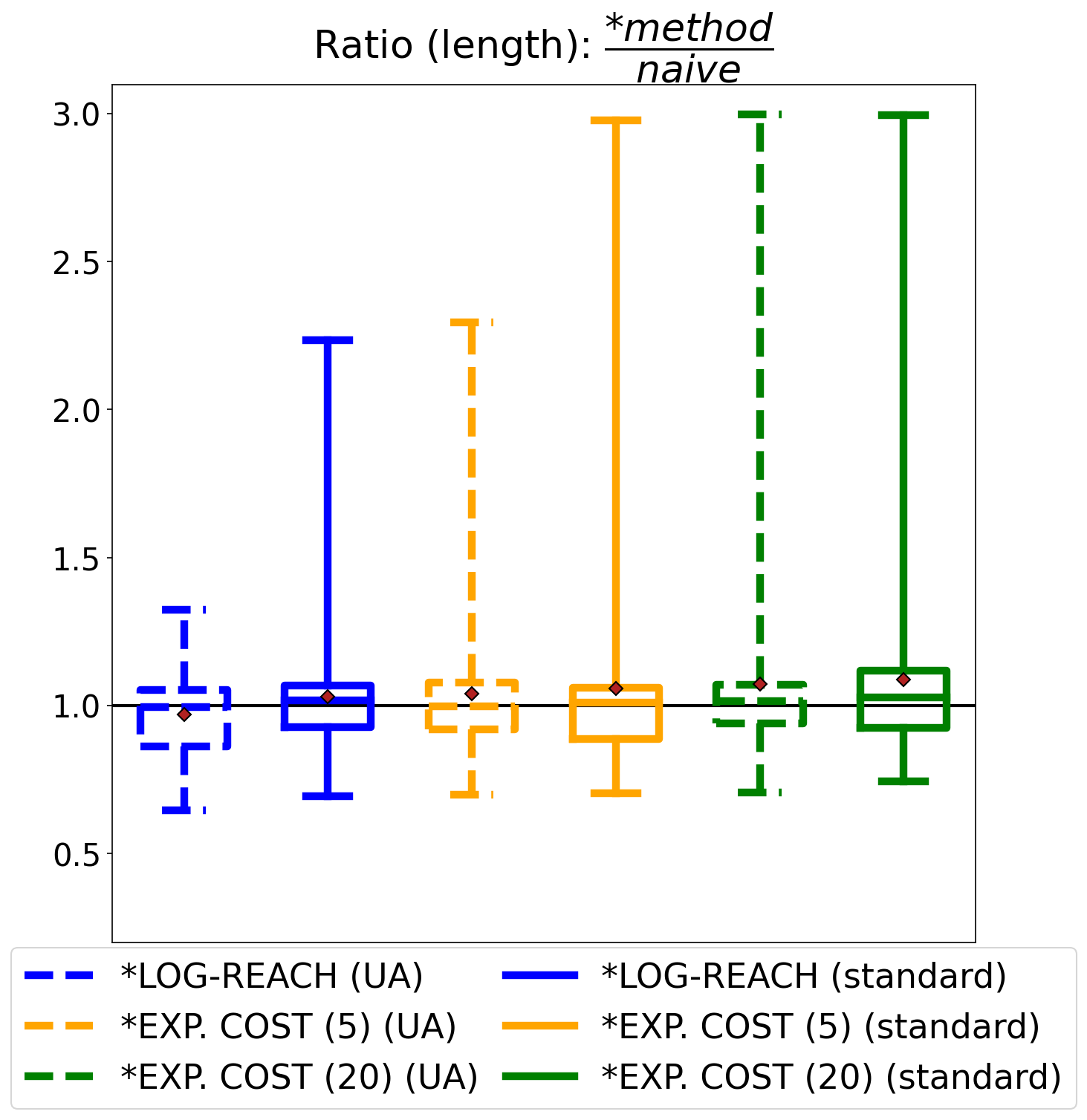}%
   }  \hspace{0.1mm}
   \subfloat[]{\includegraphics[width=.47
   \columnwidth,trim={0.5cm 0 0 1.1cm},clip, valign=t]{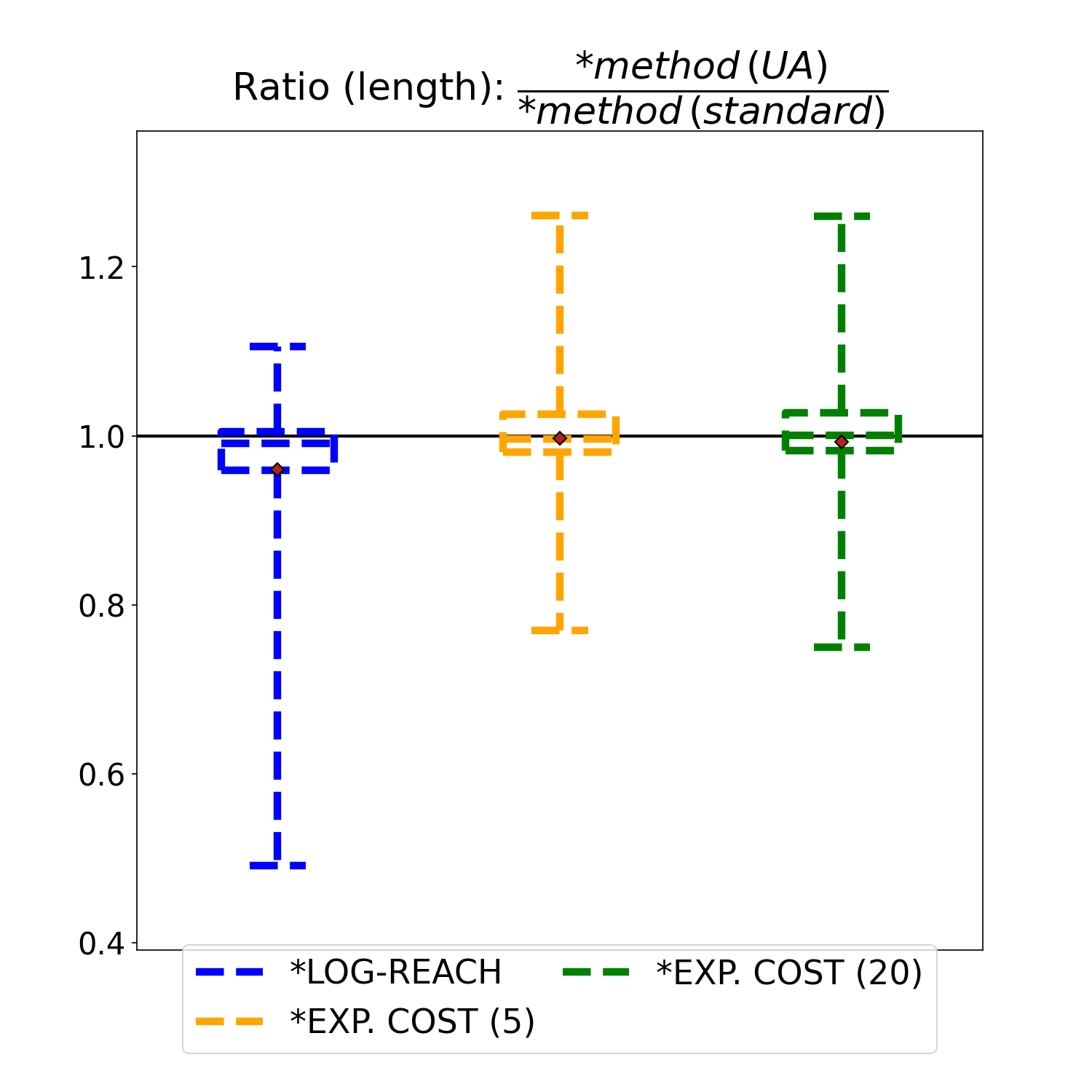}%
    }
 \caption{\small{\lucas{(a) and (b) show the planning results for the scenario with the aerial map created using a GPS-like aerial trajectory. (c) and (d) show the same comparisons for the scenario with map obtained from a simulated aerial SLAM trajectory. In all plots, path ratios below 1 (horizontal line) indicate better results. Suffixes UA and standard in the legends refer to prior maps, UA-occupancy and standard occupancy respectively, used by the navigation method. EXP.\ COST (5) and EXP.\ COST (20) refer to expected cost with parameters $C_{\text{obst}}=5$ and 20 respectively. Red markers show mean values including outliers.}}}
 \label{fig:ratio_paths}
 \end{figure}

\begin{figure*}[t]
 \centering
  \includegraphics[width=.95\textwidth]{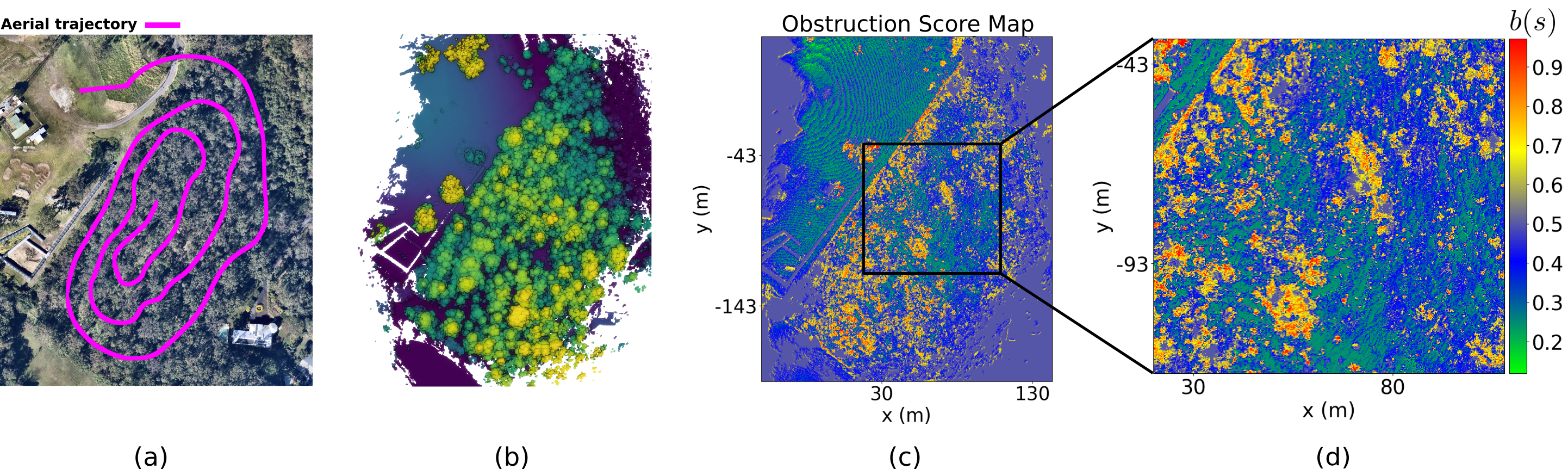}%
\caption{\small{The satellite image (a) and the aggregated aerial lidar scans (b) of the forest area used in our experiments. The resulting obstruction map generated using our proposed approach is shown in (c) along with a magnified section for better visualisation in (d). }}
\label{fig:mapping_real_data}
 \end{figure*}
We defined 40 random start and end goal pairs in the synthetic forest to be executed by our navigation method and the \lucas{na\"{\i}ve} planner. The \lucas{na\"{\i}ve} planner does not use prior maps, and is equivalent to assuming a direct path from the start to the goal as the global prior path. All 40 runs were repeated with three configurations of cost function: log-reachability \lucas{(LOG-REACH)}, and two versions of expected cost \lucas{(EXP. COST)} \eqref{eq:expected_cost} with values $C_{\text{obst}}=5$ and $20$, respectively. During online execution of the planners, we defined a 5 meter radius around the robot pose to update the maps with local information from the ground-truth environment, hence map cells in this region are changed to near 1 or 0, in case obstacles or free areas are detected, respectively, from the ground-truth map. In both methods we use D* Lite to replan in case obstacles block the robot's planned path. While our method replans on a combined map (prior map with local updates), \lucas{the na\"{\i}ve} method replans only on the map built online with local information.

The metric used for evaluation is the ratio between the executed path length of our navigation method and the \lucas{na\"{\i}ve} baseline. \lucas{We also compare the executed path length ratio between our planning using UA-occupancy map against planning using standard occupancy, to assess the benefits of our uncertainty incorporation during navigation. \figref{fig:ratio_paths} shows the planning results in the two simulated scenarios with maps generated from GPS and SLAM-like aerial trajectories. From \figref{fig:ratio_paths} (a), LOG-REACH cost, using UA and standard, both resulted in a significantly higher number of shorter paths than the \lucas{na\"{\i}ve} approach, demonstrating the advantage of using a prior aerial map for global planning and online execution}. The \lucas{EXP.\ COST}, however, showed lower performance with path costs closer to the na\"ive baseline. This performance reduction is due to some paths planned with expected cost crossing a small number of cells with high obstruction scores instead of planning longer detours, resulting in expensive replans during online execution if the planned path was blocked. Further, although the \lucas{na\"{\i}ve} approach has no prior, in cases with fewer obstructions the \lucas{na\"{\i}ve} assumption (straight-line) may actually be globally optimal or near-optimal, requiring little or no replanning. Because the log-reachability cost is more risk averse by design, it may prefer to detour even small potentially obstructed areas, yielding better results (see \figref{fig:ratio_paths} (a)). \lucas{Simulating GPS-like noise (\figref{fig:ratio_paths} (b)) did not show significant performance differences between planning with an uncertainty-aware map versus standard, confirming the intuitive fact that with an accurate aerial trajectory estimate the difference between planning with UA and standard maps is not significant.
Planning behaviour differences become more evident as uncertainty increases, as illustrated by simulations with aerial trajectories with higher noise and drift (SLAM-like). In this new scenario (SLAM-like aerial trajectory) (\figref{fig:ratio_paths} (d)), the LOG-REACH using UA maps clearly outperformed the standard counterpart, confirming that in scenarios with higher uncertainty in aerial trajectories, our mapping with pose-uncertainty can improve the planned paths. An additional consequence of the increased noise is that all methods showed reduced performance with respect to na\"ive (see \figref{fig:ratio_paths} (c)). This is expected, because as shown in \tabref{table:AUC}, higher perturbations in the aerial trajectory result in more degraded maps. However, even in this higher noise scenario, using our uncertainty-aware prior aerial map with LOG-REACH is still a better choice than na\"ive on average (see mean values shown with red markers in \figref{fig:ratio_paths}). We expect that in more complex terrains, the ground-view sensor's limited look-ahead (due to significant occlusions, which our current simulation setup does not capture) degrades the performance of na\"ive even more, ultimately risking the mission completion.    }

\section{Real-World Experiments}

We also evaluated our offline mapping and planning framework using real aerial view lidar datasets and a ground mobile robot. The airborne lidar scans were collected using a DJI M-300 quadcopter equipped with a velodyne VLP-16 lidar sensor. For ground deployment, we used a Dynamic Tracked Robot (DTR) (shown in Fig.\ref{fig:front_figure_v2}) equipped with our in-house lidar perception pack\cite{ramezani2022wildcat} for local sensing and obstacle avoidance. The tests were performed in the western forest area at the Queensland Centre for Advanced Technologies (QCAT) in Brisbane, Australia. The testing area has dense canopy cover and varied obstacles over the terrain, \eg{} high vegetation, logs, trees and narrow gaps, which makes the global navigation and local traversability highly complex. 

\begin{figure}[!b]
 \begin{center}
  \subfloat[]{\includegraphics[width=.48\columnwidth]{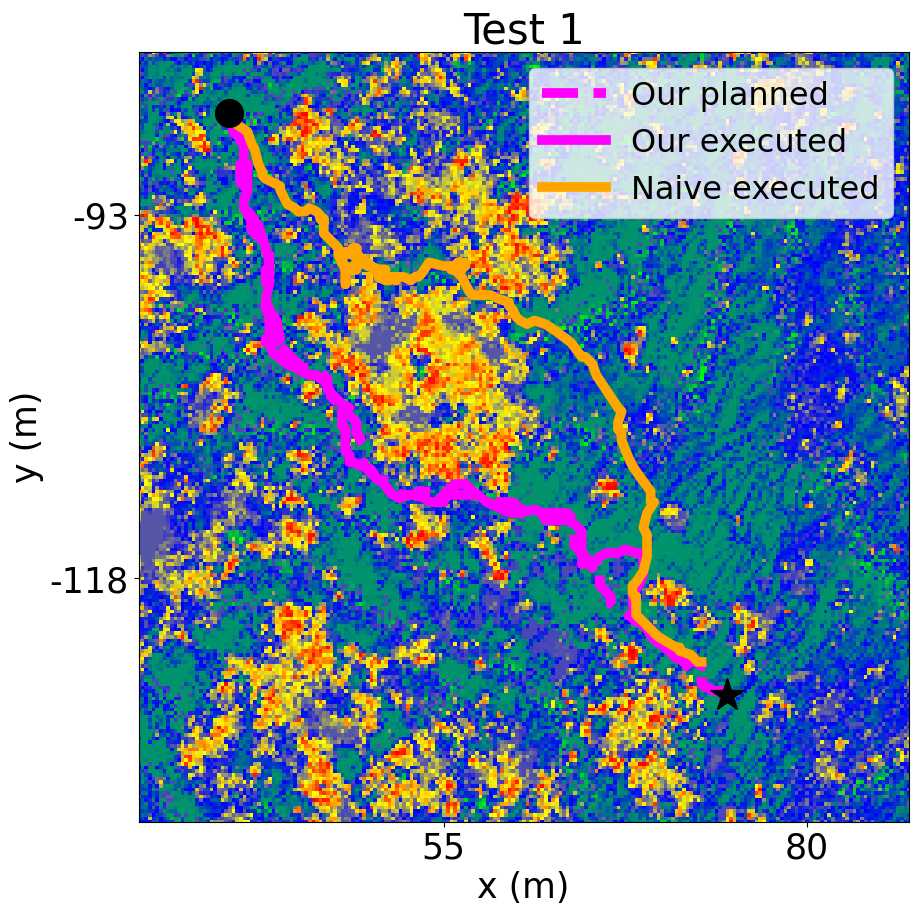}%
  } 
  \subfloat[]{\includegraphics[width=.50\columnwidth]{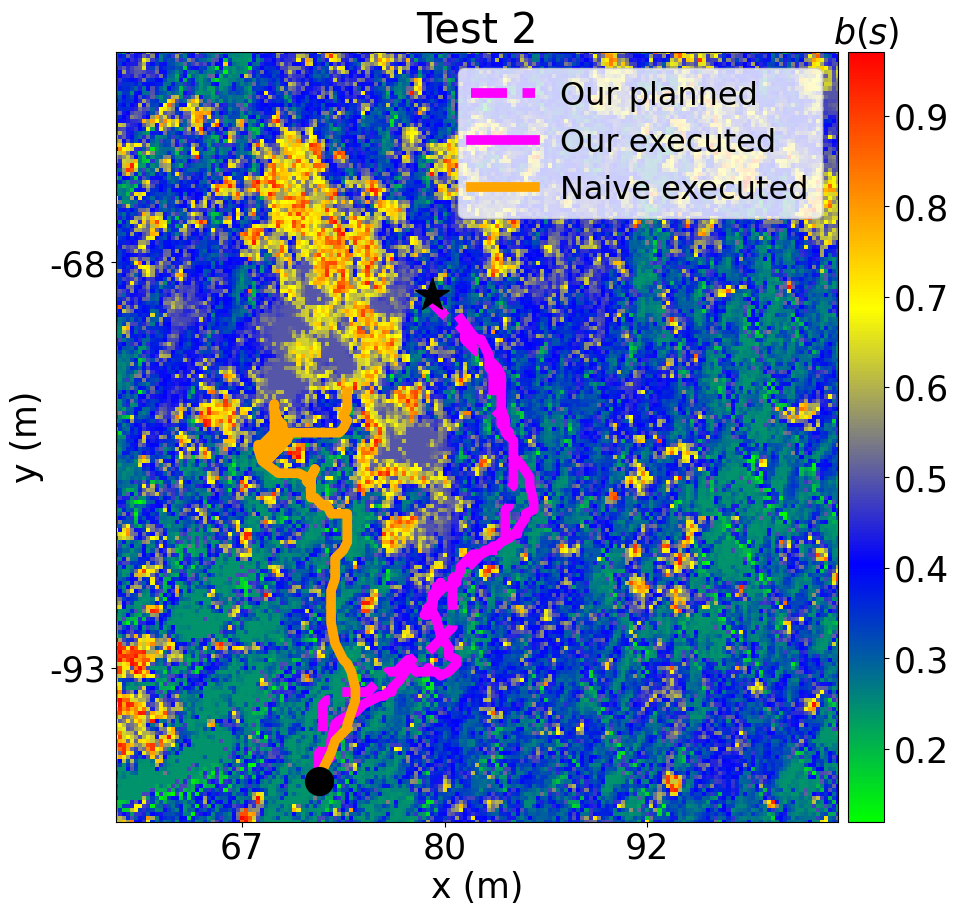}%
  }
\end{center}
\vspace{-2mm}
 \caption{\small{Real world experiments Test 1 (a) and Test 2 (b) showing the executed paths of our approach and the \lucas{na\"{\i}ve} method. The start and goal are represented by a black circle and star, respectively. In both experiments our approach guided the robot to shorter, less cluttered paths, whereas the \lucas{na\"{\i}ve} method resulted in a longer trajectory in (a) due to crossing a more cluttered area, and ultimate failure of the planner in (b) by facing a dead-end of high undergrowth.}}
 \label{fig:experiments}
 \end{figure}
 
Since it is not trivial to define ground-truth for obstructed areas in forest environments given their complexity, \eg{} cells that are visually ``obstructed" by vegetation are often traversable, \lucas{we qualitatively analysed} our mapping output from real data. %
As observed in \figref{fig:mapping_real_data} (a)-(d), even with partial canopy occlusion, our mapping using above-canopy lidar scans is capable of detecting regions at ground level with high obstruction scores (yellow to red) indicating the presence of obstacles, and free regions (green) with higher chances of traversing. Uncertain areas (blue) are also captured, providing meaningful information for planning and decision-making.

To test our full system, we defined two start and end goal pairs in the forest named Test 1 and 2. For each test location we performed two runs, one with the \lucas{na\"{\i}ve} approach where the robot used only the local perception and its embedded navigation system to reach the target without any prior information, and another run using our proposed navigation method (prior aerial mapping and global planning) with the log-reachability cost as it demonstrated superior performance in simulation. Unlike the navigation pipeline used in simulations, here our method was used only offline to create the obstruction map and plan a global trajectory from start to goal using D* Lite. For online operation, the planned trajectory was provided to the robot as a set of waypoints (transformed from the aerial map's reference frame to the robot's local frame). During online execution, the robot uses the local navigation system~\cite{Hudson_2022} with the Hybrid A* planner~\cite{HybridAStar}, and onboard perception to track the waypoints while also navigating around locally-detected obstacles. For fair comparison, the same local navigation system was used for both runs, \lucas{na\"{\i}ve} and our method; in this manner, we can assess the benefits of providing information prior to operation through a global trajectory.

During Test 1~(\figref{fig:experiments} (a)), while our approach preferred a less cluttered path on the left side of the large obstacle, the \lucas{na\"{\i}ve} approach took the right side, following a much denser path that also crossed a narrow passage among high bushes. Our method travelled a path length of $69.6$ m in comparison to $79.9$ m of the \lucas{na\"{\i}ve} trajectory, which demonstrates that our proposed navigation not only increases path efficiency, but it is also more risk averse preferring safer areas with less likelihood of having obstacles. In Test 2  (\figref{fig:experiments} (b)) the limited look-ahead of the \lucas{na\"{\i}ve} approach directed the robot to a dead-end composed of cluttered high vegetation. The local navigation was unable to detour the blocked area, and failed to reach the goal. Unlike the \lucas{na\"{\i}ve} counterpart, our method leveraged the prior information, avoiding the dead-end and taking a clear path on the right of the vegetated area to reach the goal. 
 
Although we acknowledge that the practical experiments \lucas{are not} statistically representative results, the outcomes validate the beneficial behaviours of our method observed in simulation. The field trials showed that planning using our obstruction maps in tandem with the proposed log-reachability cost leads to shorter and safer paths by avoiding cluttered areas, which ultimately increases the mission success rate.   

\lucas{Our mapping module runtime on an Intel Core i7 with 62.4GB RAM (4 CPUs) is approximately $37$ min for 28 million lidar points, sufficient for offline mapping prior to ground robot deployment. For realtime applications, GPU processing could be used to reduce the mapping runtime. }

\section{Conclusion \& Future Work}

We presented a navigation system for mapping using prior overhead lidar data and planning in complex forest environments. Our proposed mapping estimates the obstructed areas on ground level using a scoring function of occupancy values obtained from airborne lidar scans and the sensor trajectory. A Monte Carlo sampling approach to incorporate sensor pose uncertainty in occupancy maps is also presented and extensively tested in simulation, showing more accurate maps than the standard occupancy mapping. We proposed two path cost functions that take into account both obstruction uncertainty and path length. Lastly, D* Lite leverages the cost functions to plan optimal paths for ground-navigation, also allowing efficient replanning.  
Simulations and real experiments demonstrated that our navigation approach using prior information leads to shorter paths than the na\"ive baseline (no prior) and increases the chances of reaching the goal even in challenging forests scenarios. In future work we intend to learn the obstruction score map in a self-supervised fashion online \lucas{and investigate more accurate trajectory sampling techniques from the posterior of modern SLAM systems (see Sec.\ \ref{sec:expected_occ_map}).}

\balance
\bibliographystyle{IEEEtran} %
\bibliography{literature_review}

\end{document}